%% file: main.tex
\documentclass[runningheads]{llncs}

\usepackage[T1]{fontenc}

\PassOptionsToPackage{dvipsnames}{xcolor}

\usepackage{graphicx}
\usepackage{tikz}
\usetikzlibrary{positioning, calc}
\usepackage{hyperref}
\usepackage{url}
\usepackage{listings}
\usepackage{xspace}
\usepackage{array}
\usepackage{makecell}
\usepackage{graphicx}
\usepackage{multirow}
\usepackage{tcolorbox, soul} 
\usepackage{makecell, multicol, color}
\usepackage{booktabs}
\usepackage{wrapfig}
\usepackage{enumitem}
\usepackage{lipsum}

\usepackage{longtable}
\usepackage{tikz}
\usepackage[numbers]{natbib}  
\usepackage[ruled, vlined, linesnumbered, commentsnumbered, longend]{algorithm2e}
\usepackage{hyperref}

\usepackage{xcolor}
\usepackage{xspace}
\usepackage{booktabs, multirow}
\usepackage{soul}
\usepackage{listings}
\usepackage{subcaption}
\usepackage{wrapfig}
\usepackage{enumitem}
\usepackage{wrapfig}

\definecolor{codegreen}{rgb}{0,0.6,0}
\definecolor{codegray}{rgb}{0.5,0.5,0.5}
\definecolor{codepurple}{rgb}{0.58,0,0.82}
\definecolor{backcolour}{rgb}{0.97,0.97,0.95}
\definecolor{forestgreen}{rgb}{0.28,0.62,0.37}
\definecolor{codeblue}{rgb}{0,0.5,1}

\lstdefinelanguage{markdown}{
    morekeywords={*, \#, \_, `},
    sensitive=false,
    morecomment=[l]{//},   
    morestring=[b]",        
    postbreak={},
breakindent=0pt,
breakautoindent=false,
}

\lstdefinestyle{mystyle}{
    backgroundcolor=\color{backcolour},   
    commentstyle=\color{codepurple},
    keywordstyle=\color{codepurple},
    numberstyle=\tiny\color{codegray},
    stringstyle=\color{blue},
    basicstyle=\ttfamily\scriptsize,
    breakatwhitespace=false,         
    breaklines=true,                 
    captionpos=b,                    
    keepspaces=true,                 
    numbers=left,                    
    numbersep=5pt,                  
    showspaces=false,                
    showstringspaces=false,
    showtabs=false,                  
    tabsize=4
}

\usetikzlibrary{shapes, arrows}

\newif\ifextended
\extendedtrue
\ifextended
\newcommand{\extref}[1]{#1}
\else
\newcommand{\extref}[1]{#1 of the extended version of the paper~\citep{fullversion}}
\fi

\begin{document}
\input{macro}

\title{\tech: Class Invariant Synthesis using Large Language Models}

\authorrunning{C. Sun et al.}

\author{Chuyue Sun\inst{1} \and
Viraj Agashe\inst{2} \and
Saikat Chakraborty\inst{2} \and
Jubi Taneja\inst{2} \and
Clark Barrett\inst{1} \and
David Dill\inst{1} \and
Xiaokang Qiu\inst{3} \and
Shuvendu K. Lahiri\inst{2}}

\institute{Stanford University, USA\\
\and
Microsoft Research, USA\\
\and
Purdue University, USA}

\maketitle

\begin{abstract}
Formal program specifications in the form of preconditions, postconditions, and class invariants have several benefits for the construction and maintenance of programs. They not only aid in program understanding due to their unambiguous semantics but can also be enforced dynamically (or even statically when the language supports a formal verifier).  However, synthesizing high-quality specifications in an underlying programming language is limited by the expressivity of the specifications or the need to express them in a declarative manner. Prior work has demonstrated the potential of large language models (LLMs) for synthesizing high-quality method pre/postconditions for Python and Java, but does not consider class invariants.

In this work, we describe \tech, a method for co-generating executable class invariants and test inputs to produce high-quality class invariants for a mainstream language such as C++, leveraging LLMs' ability to synthesize pure functions. We demonstrate that \tech outperforms a pure LLM-based technique for generating specifications (from code) as well as prior data-driven invariant inference techniques such as Daikon. We contribute a benchmark of standard C++ data structures along with a harness that can help measure both the correctness and completeness of generated specifications using tests and mutants. We also demonstrate its applicability to real-world code by performing a case study on several classes within a widely used and high-integrity C++ codebase.

\end{abstract}

\livia{TODO: Related Work. Mention duplication in limitation}
\section{Introduction}
\input{intro}

\section{Running Example: AVL Tree}
\label{sec:avl_intro}

\dave{Where did the class invariants come from (written by the authors?)}

\input{motivation}

\section{Approach}
\label{sec:approach}
\input{approach}


\section{Results}
\label{sec:results}
\subsection{Benchmark}
\label{sec:benchmark}
\input{benchmark}
\subsection{Evaluation}
\input{experiments}

\input{results}

\section{Case study: Z3 \CodeIn{bdd\_manager} class}
\label{sec:z3}
\input{z3}

\section{Related Work}
\input{related-work}

\section{Limitations}
\label{sec:limitation}
\input{discussion}

\section{Conclusion}
\label{sec:concl}
\input{concl}

\bibliographystyle{splncs04}
\bibliography{custom}

\appendix
\include{prompt_refine}
\include{appendix/inv_table}

\include{appendix/approach_details}

\end{document}

%% file: macro.tex
\newcommand{\tech}{ClassInvGen\xspace}
\newcommand{\zthree}{Z3\xspace}

\definecolor{orange}{RGB}{255, 165, 0}

\newcommand{\cy}[1]{}
\newcommand{\livia}[1]{}
\newcommand{\shuvendu}[1]{}
\newcommand{\xiaokang}[1]{{}}
\newcommand{\viraj}[1]{{}}
\newcommand{\saikat}[1]{{}}
\newcommand{\dave}[1]{{}}
\newcommand{\clark}[1]{{}}

\newcommand{\CodeIn}[1]{{\small\texttt{#1}}}
\newcommand{\CodeInTable}[1]{{\scriptsize\texttt{#1}}}

\newcommand{\eg}{\emph{e.g.,}\xspace}
\newcommand{\ie}{\emph{i.e.,}\xspace}

\newcommand{\CodeBG}{\makebox[0pt][l]{\color{Orange!30}\rule[-0.45em]{\linewidth}{1.3em}}}

\newcommand{\CodeBGGreen}{\makebox[0pt][l]{\color{Green!30}\rule[-0.45em]{\linewidth}{1.3em}}}

\newcommand{\parabf}[1]{\noindent\textbf{#1}}

\newcommand{\CodeDelete}{\makebox[0pt][l]{\color{red!20}\rule[-0.45em]{\linewidth}{1.3em}}}
\newcommand{\CodeAdd}{\makebox[0pt][l]{\color{green!20}\rule[-0.45em]{\linewidth}{1.3em}}}

%% file: intro.tex
Invariants are predicates that hold on the program state for all executions of the program. Many invariants hold only at specific code locations.
For sequential imperative programs, it is useful to associate invariants with entry to a method (preconditions), exit from a method (postconditions), and loop headers (loop invariants).
Further, for stateful classes, class invariants are facts that hold as both preconditions and postconditions of the public methods of the class, in addition to serving as a postcondition for the class constructors for the class. 

These program invariants help make explicit the assumptions on the rest of the code, helping modular review, reasoning, and analysis. 
Program invariants are useful for several aspects of software construction and maintenance during the lifetime of a program.
First, executable program invariants can be enforced at runtime, where they provide an early indicator of state corruption, help with root causing, and allow a program to halt with an error instead of producing unexpected values. 
Runtime invariants serve as additional test oracles to amplify testing efforts to catch subtle bugs related to state corruption; this, in turn, helps with regression testing as the program evolves to satisfy new requirements. 
The utility of program invariants has led to design-by-contract in languages such as Eiffel~\cite{meyer1992eiffel}, as well as support in other languages such as Java (JML~\cite{jml}) and .NET (Code Contracts~\cite{fahndrich2010static}). 
Furthermore, for languages that support static formal verification (e.g., Dafny~\cite{leino2010dafny}, Verus~\cite{verus}, F*~\cite{swamy2011secure}, Frama-C~\cite{kirchner2015frama}), invariants can serve as a part of the specification, helping make formal verification modular and scalable. Unfortunately, invariants are underutilized because they require additional work and are sometimes difficult to write, so it would be useful to find a way to generate them automatically.

We focus specifically on automating the creation of class invariants for mainstream languages without first-class specification language support (e.g., C++) for several reasons:
\begin{itemize}
\item Class invariants are crucial for maintaining the integrity of data structures and help point to state corruption that may manifest much later within the class or in the clients.
Documenting such implicit contracts can greatly aid the understanding for maintainers of the class. 
\item Class invariants often form important parts of preconditions and postconditions for high-integrity data structures. Encapsulating such invariants and asserting them in preconditions and postconditions helps reduce bloat in the specifications. 
\item Class invariants are challenging for users to write, as writing them requires global reasoning across all the public methods for the class. 
\end{itemize}

For example, consider the class in Figure~\ref{fig:z3_dll} for a doubly-linked list, as implemented in the Z3 SMT solver.\footnote{\url{https://github.com/Z3Prover/z3/blob/master/src/util/dlist.h}}
\begin{figure}[htp]
    \centering

\begin{subfigure}[t]{0.55\linewidth}
\begin{lstlisting}[language=c++]
   void insert_before(T* other) {
        ...
        SASSERT(invariant());
        SASSERT(other->invariant());
        ...
        T* prev = this->m_prev;
        T* other_end = other->m_prev;
        prev->m_next = other;
        other->m_prev = prev;
        other_end->m_next = static_cast<T*>(this);
        this->m_prev = other_end;
        ...
        SASSERT(invariant());
        SASSERT(other->invariant());
        ...
    }
\end{lstlisting}
    
\end{subfigure}
\hfill
\begin{subfigure}[t]{0.42\linewidth}
\begin{lstlisting}[language=c++,firstnumber=17]
    bool invariant() const {
        auto* e = this;
        do {
            if (e->m_next->m_prev != e)
                return false;
            e = e->m_next;
        }
        while (e != this);
        return true;
    }
\end{lstlisting}
\end{subfigure}
    \caption{An invariant in the doubly linked list class in Z3.}
    \label{fig:z3_dll}
\end{figure}

We see that the invariant is repeated four times: as a precondition and postcondition for the object instance \CodeIn{this} and for \CodeIn{other}.
The invariant is also non-trivial, requiring local variables and a loop.

Synthesizing program invariants has been an active line of research, with both static and dynamic analysis-based approaches. 
Static analysis approaches based on variants of \textit{abstract interpretation}~\cite{cousot1977abstract} and \textit{interpolation}~\cite{henzinger2004abstractions} create invariants that are sound by construction. 
However, such techniques do not readily apply to mainstream programming languages with complex language constructs or require highly specialized methods that do not scale to large modules, since the invariants need to be additionally \textit{provably inductive} to be retained. 
On the other hand, Daikon~\cite{ernst2007daikon} and successors learn invariants dynamically by instantiating a set of templates and retaining the predicates that hold on concrete test cases. 
While applicable to any mainstream language, it is well known that Daikon-generated invariants overfit the test cases and are not sound for all test cases~\cite{polikarpova2009comparative}.
Recent works have studied fine-tuning large language models (LLMs), to learn program invariants~\citep{pei2023learning} but these methods inherit the limitations of Daikon because their training data consists of Daikon-generated invariants. More importantly, the approach has not been evaluated on stateful classes to construct class invariants. 

Recent work on \textit{prompting} LLMs such as GPT-4 to generate program invariants for mainstream languages~\cite{nl2postcond,greiner2024automated,ma2024specgen}  
has been used to generate preconditions, postconditions, and loop invariants, but these methods do not readily extend to generating class invariants. These pipelines work at single-loop or single-method scope and validate only scalar, intraprocedural predicates, so they lack the heap-aware, cross-method perspective required to state class invariants.
Further, these methods cannot construct expressive invariants that require iterating over complex data structures (such as in Figure~\ref{fig:z3_dll}) other than simple arrays. 

In this work, we introduce \tech, a novel method for generating high-quality object invariants for C++ classes through \emph{co-generation} of invariants and test inputs using LLMs such as GPT-4o. 
We leverage LLMs' ability to generate code to construct invariants that can express properties over complex data structures.
The ability to consume not only the code of a class but also the surrounding comments and variable names helps establish relationships difficult for purely symbolic methods.
Since an LLM can generate incorrect invariants, the method also generates test inputs to \emph{heuristically} prune incorrect candidate invariants. 

We leverage the framework proposed by Endres et al.~\cite{nl2postcond} to evaluate the test-set correctness and completeness given a set of hidden validation tests and mutants. 
We contribute a new benchmark comprising standard C++ data structures along with a harness that can help measure both the correctness and completeness of generated invariants (Section~\ref{sec:benchmark}).
We demonstrate that \tech outperforms a pure LLM-based technique for generating program invariants from code (Sections~\ref{subsec:correctness}--\ref{subsec:completeness}) as well as prior data-driven invariant inference techniques such as Daikon (Section~\ref{subsec:daikon_compare}).
We also demonstrate its applicability for real-world code by performing a case study on a set of classes in the Z3 SMT solver codebase, including the relatively complex \CodeIn{bdd\_manager} class; the developers of the codebase confirmed most of the new invariants proposed by \tech for these modules (Section~\ref{sec:z3}). 

Our contributions are summarized below:
\begin{itemize}
    \item We introduce a new technique for invariant-test co-generation by combining simple static analysis with LLMs and implement an end-to-end prototype (Section~\ref{sec:approach}).
    \item We introduce a high-quality \tech-instrumented benchmark for evaluating object invariants  (Section~\ref{sec:benchmark}).
    \item We investigate LLM-assisted class invariant synthesis (Sections~\ref{subsec:correctness}--\ref{subsec:daikon_compare}).
    \item We conduct a case study on Z3 class modules using \tech (Section~\ref{sec:z3}).
\end{itemize}

\tech is conceived as a \emph{specification-drafting aid}: it produces candidate invariants that a developer can accept, refine, or discard, thereby following the long-advocated “human-in-the-loop’’ paradigm in specification mining\,\cite{newcomb2019usinghumanintheloopsynthesisauthor}. 
Our aim is not to replace expert judgement, but to accelerate it.

%% file: motivation.tex

Throughout this paper, \CodeIn{AvlTree}~\cite{avltree,cpp_ds} from our benchmark (Section~\ref{sec:benchmark}) will be used as a running example to illustrate the workflow. An AVL tree is a self-balancing binary search tree (BST) where the difference in heights between the left and right subtrees of any node (called the balance factor) is at most 1. This ensures that the tree remains approximately balanced. In this implementation shown in Figure~\ref{fig:avl_header} in Section~\ref{subsec:generation}, the \CodeIn{AvlTree} class contains several public methods: \CodeIn{insert}, \CodeIn{remove}, \CodeIn{contains}, \CodeIn{clear}, \CodeIn{height}, \CodeIn{size}, \CodeIn{empty}, \CodeIn{in\_order\_traversal}, \CodeIn{pre\_order\_traversal}, \CodeIn{post\_order\_traversal}. \CodeIn{AvlTree} maintains several class invariants determined by the authors of this paper: 
\begin{itemize}
    \item BST Property: Left child values are less than the node's value, and right child values are greater.
    \item Balance Factor: For each node, the difference in the heights of the left and right subtrees is between -1 and 1.
    \item Correct Heights: The height of each node is 1 plus the maximum height of its children.
\end{itemize}

Each of these invariants should hold true before and after every public method call, and be established after the constructor method. The task is to infer these high-quality invariants from the source code.

%% file: approach.tex
\begin{figure*}[ht]
    \centering
    \includegraphics[width=0.95\textwidth]{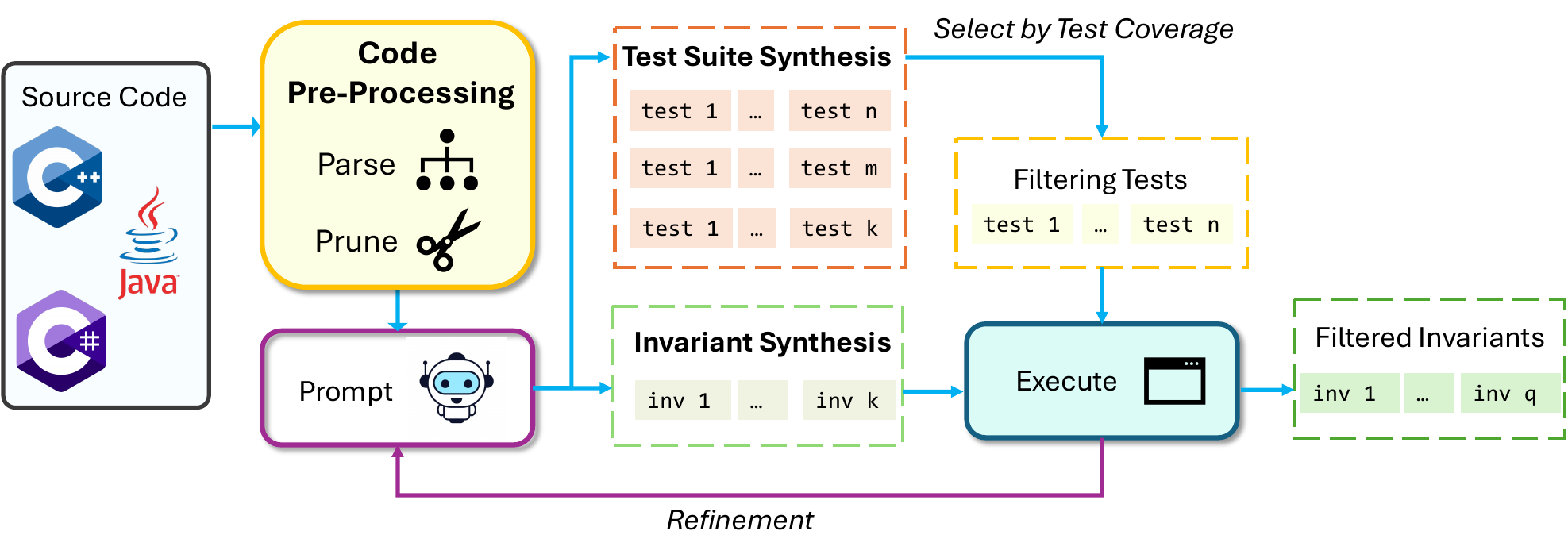}  
    \caption{Overview of \tech.}
    \label{fig:framework}
\end{figure*}


An overview of the \tech framework is shown in Figure~\ref{fig:framework}. It outlines an automated pipeline for inferring class invariants from source code. 
\tech takes a complete source program as input and outputs invariant candidates it has identified with high confidence (called \textit{filtered invariants}).
\tech starts with a preprocessing step which performs static analysis on the program (Section~\ref{subsec:generation}).  Next, an LLM is used to generate candidate invariants and filtering test suites (Section~\ref{subsec:generation}).  Then, the code is instrumented to facilitate checking candidate invariants (Section~\ref{subsec:generation}).  
Finally, \tech uses generated tests to prune invariant candidates (Section~\ref{sec:testing}), and a refinement loop is used to iteratively improve the results (Section~\ref{sec:refinement}). 
\shuvendu{Still cannot justify the test ranking}


\subsection{Generation}
\label{subsec:generation}

\subsubsection{Preprocessing}
\label{sec:preprocess}

As illustrated in Figure~\ref{fig:framework}, the generation phase begins with a static analysis of the source program. \tech uses a Tree-Sitter-based parser for program preprocessing; Tree-Sitter~\cite{treesitter} is a parser generator tool that constructs a syntax tree from source files.

\tech parses the entire source program into an abstract syntax tree (AST) to extract class members and their recursive dependencies. It then identifies the target class and gathers details (e.g., method declarations, field declarations, and subclass definitions) relevant to forming correct class invariants. \tech recursively analyzes all identified classes (i.e., the target class and its subclasses) and performs a topological sort to prepare generation from the leaf class upward as shown in Algorithm~\ref{algo:algo}.
\shuvendu{If we define concepts such as get\_method\_bodies, we have to show how we use them.}

\begin{algorithm}[t]
\caption{Function to generate invariants for target class AST}
\small
\label{algo:algo}
\DontPrintSemicolon
\SetKwProg{Fn}{Function}{:}{}

\SetKwFunction{GenerateInvariant}{\textsc{GenerateInvariant}}
\SetKwFunction{getClassRecursively}{\textsc{getClassRecursively}}
\SetKwFunction{sort}{\textsc{reverseToplogicalSort}}
\SetKwFunction{getClassDependencies}{\textsc{getClassDependencies}}
\SetKwFunction{needsInvariant}{\textsc{needsInvariant}}
\SetKwFunction{getCodeForClass}{\textsc{getCodeForClass}}
\SetKwFunction{getCompletions}{\textsc{generateInvariantWithLLM}}

\SetKwFunction{getClassMethods}{\textsc{getClassMethods}}
\SetKwFunction{getText}{\textsc{getText}}

\Fn{\GenerateInvariant{target\_class}}{
    \If{target\_class.id $\in$ invariants\_dict}{
        \Return{invariants\_dict[target\_class.id]}\; \label{line:cache}
    }
    
    dep\_classes $\leftarrow$ \getClassRecursively{target\_class}\; \label{line:collect}
    rev\_topsorted\_classes $\leftarrow$ \sort{dep\_classes}\; \label{line:sort}
    
    \ForEach{class $\in$ rev\_topsorted\_classes}{
            class\_code $\leftarrow$ \getCodeForClass{class, invariants\_dict}\; \label{line:classcode}
            
            \ForEach{dep $\in$ \getClassDependencies{class}}{
            \tcp{Invariant: invariants have been generated for dep}
                dep\_code $\leftarrow$ \getCodeForClass{dep, invariants\_dict}\; \label{line:dev-code}
                class\_code $\leftarrow$ class\_code  + dep\_code\; \label{line:dev-code-ed}
            }
            
            invariants\_dict[class.id] $\leftarrow$ \getCompletions{class\_code}\; \label{line:geninv}
    }
    
    \Return{invariants\_dict[target\_class.id]}\;
}

\Fn{\getCodeForClass{class, invariants\_dict}}{
    class\_text $\leftarrow$ class.get\_declaration\_text()\tcp*{Header} \label{line:helper-1}
    \If{class.id $\in$ invariants\_dict \textbf{ and } invariants\_dict[class.id]}{
        \tcp{Get generated invariants}
        class\_text $\leftarrow$ invariants\_dict[class.id] + class\_text \label{line:helper-2}
    }
    
    \Return{class\_text}\;
}
\end{algorithm}


\subsubsection{Generation by LLM}
\label{sec:llm}

After building the source program AST, \tech uses LLMs to analyze the class module and infers both invariants for the target class and tests that exercise the class's implementation as thoroughly as possible. \tech uses a fixed system prompt that defines class invariants and outlines two main tasks: (1) generating class invariants from the source code, and (2) creating a test suite of valid API calls without specifying expected outputs (see prompt details in \extref{the Appendix~\ref{appendix:prompt}}).
Next, \tech instantiates a user prompt template with the actual target class.

From the source program AST, \tech identifies program dependencies and populates the prompt template with the leaf struct/class. Starting from the leaf nodes, \tech leverages previously generated invariants by including them in the prompt for later classes. 
\shuvendu{I am still unconvinced that class invariants need to be exposed to clients without expanding to pre/post conditions.}
To accommodate the LLM's context window limit, only the relevant child classes of the current target class are included in the prompt, with method implementations and private fields/methods hidden when necessary. An algorithm for this process is presented in Algorithm~\ref{algo:algo}. Algorithm~\ref{algo:algo} presents the invariant generation process for a source program AST. The main function \GenerateInvariant takes a \CodeIn{target\_class} and leverages a caching mechanism through \CodeIn{invariants\_dict} to avoid redundant computations (Line~\ref{line:cache}).The algorithm first collects dependent classes via \getClassRecursively (Line~\ref{line:collect}) and sorts them using \sort to ensure dependency-aware processing (Line~\ref{line:sort}). For each \CodeIn{class} in the sorted order, it constructs the necessary context by obtaining the class code through \getCodeForClass (Line~\ref{line:classcode}). For each dependency \CodeIn{dep} of the current \CodeIn{class}, it retrieves the \CodeIn{dep\_code} and concatenates it with \CodeIn{class\_code} (Lines~\ref{line:dev-code}--\ref{line:dev-code-ed}). 

The algorithm then generates invariants using \getCompletions and stores them in \CodeIn{invariants\_dict} (Line~\ref{line:geninv}).

The helper function \getCodeForClass constructs class representations by combining the declaration text with any existing invariants from \CodeIn{invariants\_dict} (Lines~\ref{line:helper-1}--\ref{line:helper-2}).
The algorithm concludes by returning \CodeIn{final\_invariants} for the target class, effectively managing the invariant generation process while respecting LLM context limitations.

\tech accommodates large codebases by dividing the source program into smaller modules that fit within the LLM's context window. It then iteratively generates invariants and test cases, starting from leaf classes and working up towards the root class. At each step, \tech leverages previous invariants generated for child classes to inform the invariants for parent classes.

For the \CodeIn{AvlTree} example, we begin by instantiating the prompt with \CodeIn{Node} for annotation, followed by \CodeIn{AvlTree}, since \CodeIn{Node} is a subclass of \CodeIn{AvlTree}, as illustrated in Figure~\ref{fig:avl_header}. In this specific case, however, Algorithm~\ref{algo:algo} does not make a difference due to the small size of the source program; the entire \CodeIn{AvlTree} code fits within the LLM's context window easily.

\begin{figure}[t]
    \centering
\begin{lstlisting}[language=c++, escapechar=!, basicstyle=\ttfamily\scriptsize, basewidth=0.5em]
class AvlTree {
public:
  AvlTree();
  AvlTree(const AvlTree &t);
  AvlTree &operator=(const AvlTree &t);
  ~AvlTree();

  void insert(const T &v);
  void remove(const T &v);
  bool contains(const T &v);
  void clear();
  int height();
  int size();
  bool empty();
  std::vector<T> in_order_traversal() const;
  std::vector<T> pre_order_traversal() const;
  std::vector<T> post_order_traversal() const;
private:
  struct Node {
    T data;
    std::unique_ptr<Node> left;
    std::unique_ptr<Node> right;
    int balance_factor();
  };
// rest of the file
}
\end{lstlisting}
    \caption{Header file of AvlTree.}
    \label{fig:avl_header}
\end{figure}

In contrast, when working with the \CodeIn{bdd\_manager} class in Z3 (around 1700 lines of code), \tech begins generation with \CodeIn{bdd}, a subclass of \CodeIn{bdd\_manager}. Algorithm~\ref{algo:algo} enables \tech to partition the \CodeIn{bdd\_manager} class and outputs meaningful class invariants (see Section~\ref{sec:z3}).


\subsubsection{Instrumentation}
\label{sec:instrumentation}

To check candidate invariants, each public method is automatically instrumented with a \CodeIn{check\_invariant} call at both the start and end of its implementation. This allows us to verify that invariants hold both before and after method execution. Each invariant is implemented as a method call to prevent conflicts with local variables.

When a specific invariant is being checked, its code is plugged into the \CodeIn{check\_invariant} function with assertions. This ensures that during pruning, whenever a public API call is made, each invariant candidate is automatically verified.

Additional examples of instrumentation and invariant checking are provided in \extref{Appendix~\ref{sec:appendix_approach_details}}.

\subsection{Heuristic Pruning}
\label{sec:testing}

The LLM generates test suites that serve as filters for invariant candidates. To select the most effective test suite, we use line coverage as a metric, as it provides a straightforward proxy for test suite completeness. The test suite with the highest coverage becomes our set of \emph{filtering tests}.

When generating tests, \tech creates valid sequences of API calls without asserting expected behavior, since our goal is to filter invariant candidates rather than test the source program directly. Among all generated test suites, we compile and run each one with the source program, selecting the one with the highest line coverage as the \textit{filtering tests}.

\tech dynamically expands the \textit{filtering tests} only if coverage falls below a specified threshold (default $80\%$). In our experiments, each benchmark task's suite of \textit{filtering tests} includes 5 to 15 individual tests, with each test comprising 5 to 20 lines of code (see example in \extref{Appendix~\ref{appendix:prompt}}).

If an invariant candidate successfully compiles and runs with the \textit{filtering tests}, it is designated a \textit{filtered invariant} and included in the final output of \tech.

\subsection{Refinement}
\label{sec:refinement}

For invariants that fail during compilation or runtime, \tech implements a feedback-driven refinement process. For each failing invariant, the system collects compiler output, error messages, and test results, and then feeds this information back to the LLM using a dedicated prompt template. 

This feedback loop allows \tech to repair failing invariants by providing the LLM with specific error information and the context in which the error occurred. We set a default threshold of 3 refinement attempts per invariant, balancing the cost of LLM calls with the benefit of repairs.

Refinement allows \tech to fix common issues such as type errors, undefined references, and logical inconsistencies. More detailed examples of the refinement process are provided in \extref{Appendix~\ref{sec:appendix_approach_details}}.



%% file: benchmark.tex

\begin{table}[!htp]\centering
\caption{Characteristics of the benchmark data structures. \# LoC represents lines of code, \# methods indicates the number of implemented methods, and \# dep. shows the number of dependent classes for each data structure.}\label{tab:benchmark-detail}
\small
\begin{tabular}{@{}l@{\hspace{8pt}}|rrrrrrrrr@{}}\toprule
&\multicolumn{1}{c}{avl\_} &\multicolumn{1}{c}{binary\_} &\multicolumn{1}{c}{hash\_} &\multicolumn{1}{c}{heap} &\multicolumn{1}{c}{linked\_} &\multicolumn{1}{c}{queue} &\multicolumn{1}{c}{red\_black\_} &\multicolumn{1}{c}{stack} &\multicolumn{1}{c}{vector} \\
&\multicolumn{1}{c}{tree} &\multicolumn{1}{c}{search\_tree} &\multicolumn{1}{c}{table} & &\multicolumn{1}{c}{list} & &\multicolumn{1}{c}{tree} & & \\\midrule
\# LoC &249 &229 &176 &135 &172 &117 &282 &96 &117 \\ 
\# methods &25 &22 &11 &14 &14 &12 &25 &11 &18 \\
\# dep. &1 &1 &0 &0 &0 &0 &1 &0 &0 \\
\bottomrule
\end{tabular}
\end{table}

We use a C++ implementation of classical data structures for our micro-benchmarks~\cite{cpp_ds}, which include 9 data structures with associated unit tests: \CodeIn{AvlTree}, \CodeIn{BinarySearchTree}, \CodeIn{HashTable}, \CodeIn{Heap}, \CodeIn{LinkedList}, \CodeIn{Queue}, \CodeIn{RedBlackTree}, \CodeIn{Stack}, and \CodeIn{Vector}.
Table~\ref{tab:benchmark-detail} shows the statistics of these benchmark data structures.
To ensure correctness, we thoroughly examined each benchmark example and corrected a few implementation bugs, treating this refined benchmark as the ground truth. All subsequent experiments are based on this benchmark setup.
\shuvendu{It would be interesting to mention how the bug was found with the generated invariants and tests, with the user in the loop to not remove an invariant that failed the test. Perhaps informally, in the Discussion section.}


In addition to textbook examples, we also conducted experiments on utility classes~\cite{z3util} from \zthree~\cite{z3}, including \CodeIn{ema}, \CodeIn{dlist}, \CodeIn{heap}, \CodeIn{hashtable}, \CodeIn{permutation}, \CodeIn{scoped\_vector}, and the most complex class, \CodeIn{bdd\_manager}. The latter will be discussed in detail as a case study, including an evaluation with one of the authors of \zthree.


%% file: experiments.tex

In this section, we evaluate the quality of \tech invariants. 
Specifically, we explore the following research questions:
\begin{enumerate}
    \item How many of these invariants are \textbf{correct} (with respect to the user-provided test cases), and do they capture essential properties of the source code (Section~\ref{subsec:correctness})?
    \item How \textbf{complete} are the invariants in their ability to distinguish the correct program from buggy counterparts (Section~\ref{subsec:completeness})? 
    \item How does \tech compare to a state-of-the-art technique in invariant generation (namely Daikon, the most widely adopted tool for dynamic invariant synthesis) (Section~\ref{subsec:daikon_compare})?
\end{enumerate}

Our experiments were conducted on a machine with 24 CPU cores and 64 GB of RAM. We implemented \tech using GPT-4o as the underlying LLM, with its default temperature setting of 1.



%% file: results.tex
\begin{wrapfigure}{r}{0.5\textwidth}
    \centering
    \includegraphics[width=0.5\textwidth]{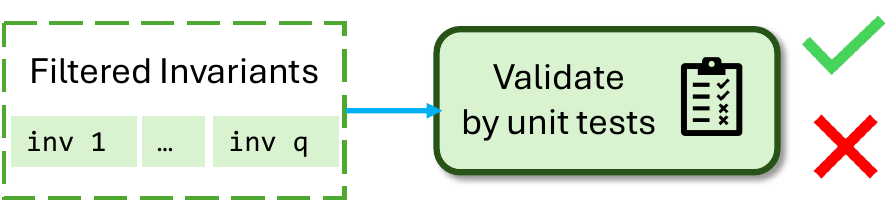}
    \caption{Evaluation of \tech generated invariants}
    \label{fig:validation}
\end{wrapfigure}

\subsection{Correctness}
\label{subsec:correctness}

\tech produces \textit{filtered invariants}, which we evaluate using an automated pipeline (Figure~\ref{fig:validation}) against our benchmark (Section~\ref{sec:benchmark}). A \textit{filtered invariant} is considered correct if it reports no errors for any tests that successfully compile and run. Our manual review confirmed that all validated \textit{filtered invariants} are indeed correct, capturing essential properties of the data structures.

\parabf{Invariant-Only Generation.}
Table~\ref{ds_0shot} shows results.
When generating invariants in isolation, \tech produces an average of $25$ unique invariants per benchmark (the number is essentially the same after filtering those that don't compile) with a $77\%$ pass rate against unit tests.

\parabf{Benefits of Co-Generation and Refinement.}
Table~\ref{ds_refine} shows results with test co-generation.  Filtering using tests reduces the average number of invariants to 17, but all are correct.
After refinement, the number of \textit{filtered invariants} grows from $17$ to $22$ per example, representing a $29\%$ increase. This demonstrates \tech's ability to transform potentially buggy invariants into valid ones through feedback-guided refinement.

\parabf{Summary.}
\tech's invariant-test co-generation approach improves correctness from $77\%$ to $100\%$. The \textit{filtering tests} effectively identify valid invariants, while the refinement process successfully expands the set of correct invariants.

\input{ablation_tables}

\subsection{Completeness}
\label{subsec:completeness}

\begin{table}[h!]
    \centering
    \scriptsize
    \setlength{\tabcolsep}{3pt}
    \caption{\tech The table shows the number of mutants that survive unit tests alone (Unsolved Base) as well as the number of additional mutants killed by \tech.}
    \begin{tabular}{lrrr}
        \toprule
        \textbf{Data Structure} & \textbf{Unsolved Base (\#(\%))} & \textbf{Add. by \tech (\#)} & \textbf{Impr. (\%)} \\
        \midrule
        binary\_search\_tree & 107(23.67) & 7 & 6.54 \\ 
        hash\_table & 258(37.83) & 38 & 14.73 \\ 
        heap & 108(32.24) & 12 & 11.11 \\ 
        linked\_list & 57(13.54) & 2 & 3.51 \\ 
        red\_black\_tree & 184(27.10) & 9 & 4.89 \\ 
        stack & 67(28.39) & 6 & 8.96 \\ 
        vector & 101(29.79) & 33 & 32.67 \\ 
        avl\_tree & 84(17.57) & 0 & 0.00 \\ 
        queue & 91(26.30) & 11 & 12.09 \\
        \midrule
        \textbf{Total} & \textbf{1057} & \textbf{118} & \textbf{11.16} \\
        \bottomrule
    \end{tabular}
    \label{table:specbot_baseline_comparison}
\end{table}

To evaluate completeness, we use mutation testing.
We generate mutants using mutate\_cpp~\cite{mutatecpp}, producing between 236 and 682 mutants per program. We focus on mutants that either compile successfully but crash during execution or survive execution without errors.

\begin{figure}[t]
    \centering
    \includegraphics[width=0.95\textwidth]{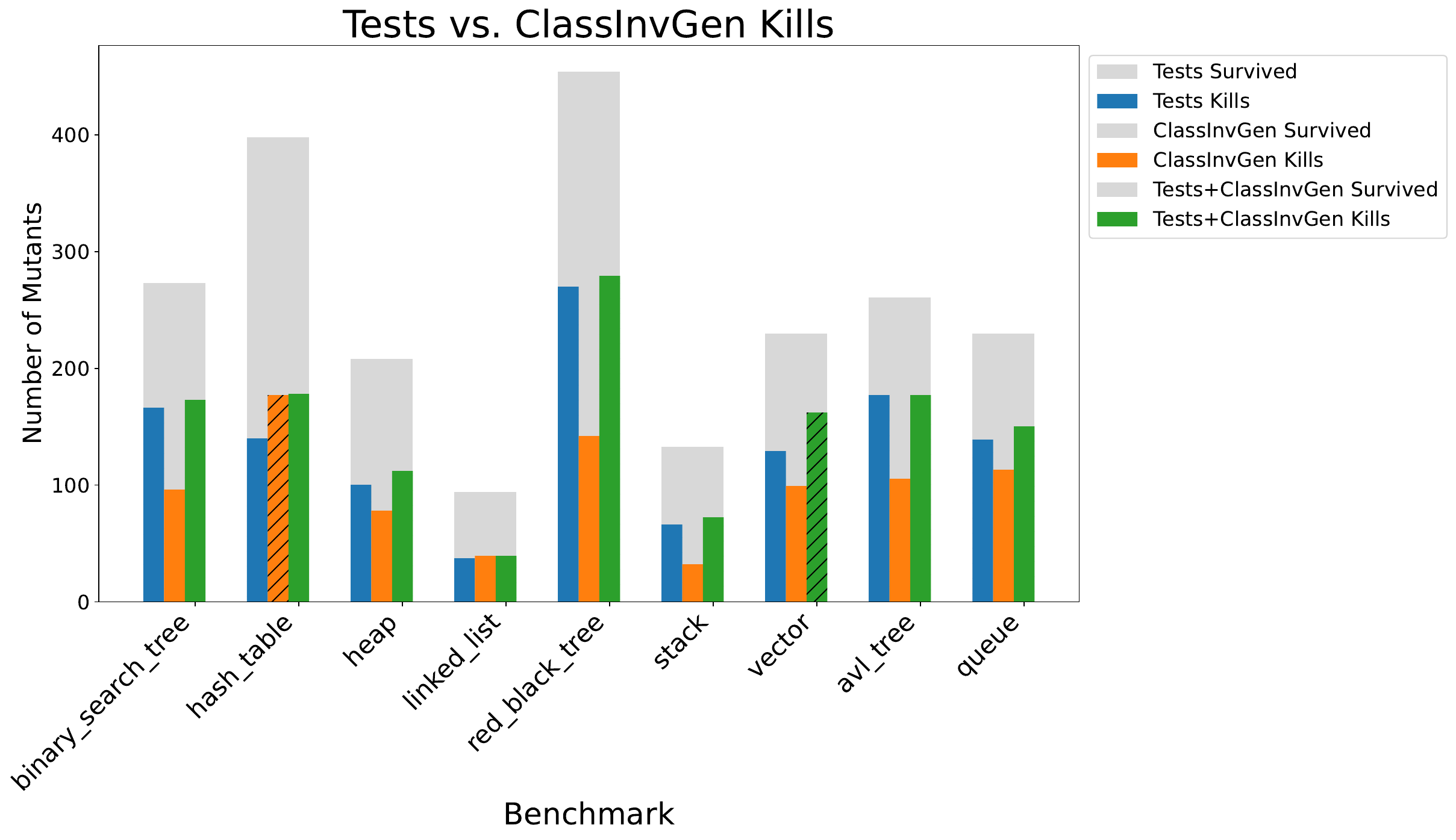}
    \caption{Completeness Experiment Result. The 3 bars from left to right are Tests, \tech, Tests+\tech. Tests+\tech kills the most mutants.}
    \label{fig:mutation}
\end{figure}

We conducted experiments to see how many mutants are killed (i.e., identified as incorrect) using three configurations: unit tests only, \tech only, and unit tests with \tech (strongest test oracles). As shown in Table~\ref{table:specbot_baseline_comparison}, \tech's invariants identify an additional $11.2\%$ of mutants on average compared to unit tests alone, with improvements reaching up to $32.67\%$ for specific data structures.

Figure~\ref{fig:mutation} shows how many mutants are killed with each configuration for each benchmark.  Combining tests with \tech kills the most mutants. Figures~\ref{fig:inv_addition_kill} and \ref{fig:inv_addition_kill_const} show examples of mutants that survived unit tests but were killed by \tech invariants.

\begin{figure}[h]
\centering
\begin{lstlisting}[language=c++, escapechar=!]
 void HashTable::clear_table() {
   this->table.clear();
   this->_num_elements = 0;
!\CodeDelete\textbf{-}!  this->_size = 0;
!\CodeAdd\textbf{+}!  this->_size += 0;
 }
}
\end{lstlisting}
    \caption{Mutant that survived unit tests but was killed by \tech}
    \label{fig:inv_addition_kill}
\end{figure}

 \begin{figure}[h]
\centering
\begin{lstlisting}[language=c++, escapechar=!]
   this->hash_function = hash_function;
   this->_num_elements = 0;
   this->_size = size;
!\CodeDelete\textbf{-}!  this->load_factor = 0.75;
!\CodeAdd\textbf{+}!  this->load_factor = -0.75;
   this->table =
       std::vector<std::shared_ptr<std::vector<std::pair<Key, Value>>>>(size);
 }
\end{lstlisting}
    \caption{Another Mutant that survived unit tests but was killed by \tech}
    \label{fig:inv_addition_kill_const}
\end{figure}

\subsection{Comparison of \tech vs. Daikon}
\label{subsec:daikon_compare}

We compared \tech with Daikon~\cite{ernst2007daikon}, using \textit{filtering tests} to generate program traces for Daikon's invariant detector. On average, each benchmark example has around 5 Daikon invariants, with some being incorrect (see summary in Table~\ref{tab:daikon_invariants} and complete results of invariants in \extref{Appendix~\ref{appendix:daikon}}).

\begin{table}[h!]
    \centering
    \small
    \caption{Daikon Incorrect Invariants per Benchmark}
    \begin{tabular}{lcc}
        \toprule
        \textbf{Data Structure} & \textbf{Total \# Invariants} & \textbf{ Incorrect Invariants} \\
        \midrule
        hash\_table           & 8 & 1 \\
        binary\_search\_tree  & 3 & 0 \\
        heap                  & 10 & 1 \\
        red\_black\_tree      & 2 & 0 \\
        avl\_tree             & 4 & 0 \\
        vector                & 3 & 1 \\
        stack                 & 6 & 2 \\
        queue                 & 7 & 1 \\
        linked\_list          & 4 & 1 \\
        \midrule
        \textbf{Average}    & \textbf{5.2} & \textbf{0.78} \\
        \bottomrule
    \end{tabular}
    \label{tab:daikon_invariants}
\end{table}

Through manual review, we identified 7 incorrect Daikon invariants that pass unit test validation. These invariants pass because both the \textit{filtering tests} and unit tests coincidentally constructed similar data structures.

\begin{figure}[t]
    \centering
    \includegraphics[width=0.95\textwidth]{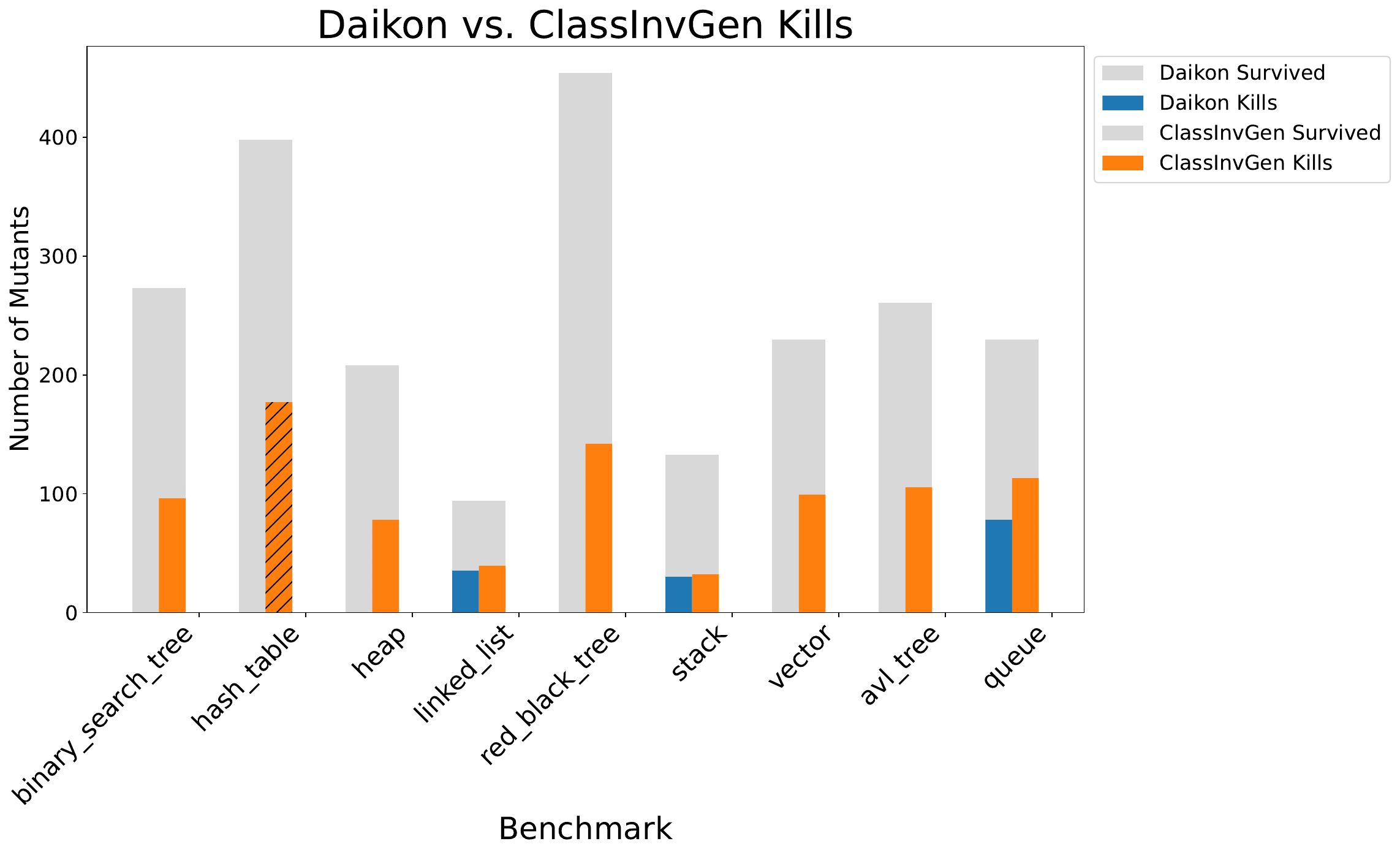}
    \caption{Daikon vs. \tech Kills}
    \label{fig:daikon_specbot}
\end{figure}

Most Daikon invariants simply indicate that class pointers are not null (27 of 40 correct invariants) or that element counts are non-negative (6 invariants). The most valuable invariants, like \CodeIn{this->n < this->maxSize} in \CodeIn{Stack}, have more impact on identifying mutants (Figure~\ref{fig:daikon_specbot}).

This shows a key weakness of Daikon: it cannot differentiate between universally true invariants and those that hold only in specific test contexts. LLMs are better at capturing true ``class'' invariants that are inherent to the data structure rather than incidental to the tests.

%% file: ablation_tables.tex
\begin{table}[ht]
\centering
\small
\setlength{\tabcolsep}{4pt}  
\begin{tabular}{lcccr}
\toprule
\textbf{Example}           & \textbf{\# inv.} & \textbf{\# compiles} & \textbf{\# pass tests} & \textbf{pass rate} \\ 
\midrule
\textbf{avl\_tree}          & 30            & 28                  & 17                    & 56.7\%             \\ 
\textbf{queue}              & 30            & 30                  & 30                    & 100.0\%            \\ 
\textbf{linked\_list}       & 32            & 32                  & 24                    & 75.0\%             \\ 
\textbf{binary\_search\_tree} & 25            & 25                  & 24                    & 96.0\%             \\ 
\textbf{hash\_table}        & 29            & 29                  & 26                    & 89.7\%             \\ 
\textbf{heap}               & 22            & 22                  & 12                    & 54.5\%             \\ 
\textbf{red\_black\_tree}   & 26            & 26                  & 15                    & 57.7\%             \\ 
\textbf{stack}              & 15            & 15                  & 11                    & 73.3\%             \\ 
\textbf{vector}             & 18            & 18                  & 16                    & 88.9\%             \\ 
\midrule
\textbf{Average}            & 25.22         & 24.56               & 19.44                 & 77.0\%             \\ 
\bottomrule
\end{tabular}
\caption{Invariant-only results from 9 benchmarks show an average of 25 invariants per benchmark, with $77\%$ passing unit tests.}
\label{ds_0shot}
\end{table}

\begin{table}[ht]
\centering
\scriptsize  
\setlength{\tabcolsep}{3pt}  
\begin{tabular}{lcccc}
\toprule
\textbf{Example} & \textbf{\# filter tests} & \textbf{\# filtered inv.} & \textbf{\# good inv.} & \textbf{pass rate} \\ 
                 & \textbf{(coverage)}      &                           & \textbf{(1 refine)}   & \textbf{(1 refine)}  \\ 
\midrule
\textbf{avl\_tree}           & 10 (91.7\%)   & 15          & 20  & 100.0\% \\ 
\textbf{queue}               & 8 (100.0\%)   & 20          & 29  & 100.0\% \\ 
\textbf{linked\_list}        & 8 (92.3\%)    & 22          & 36  & 100.0\% \\ 
\textbf{binary\_search\_tree} & 12 (95.6\%)  & 16          & 19  & 100.0\% \\ 
\textbf{hash\_table}         & 9 (90.3\%)    & 25          & 27  & 100.0\% \\ 
\textbf{heap}                & 7 (96.0\%)    & 10          & 11  & 100.0\% \\ 
\textbf{red\_black\_tree}    & 13 (85.4\%)   & 11          & 18  & 100.0\% \\ 
\textbf{stack}               & 8 (100.0\%)   & 14          & 14  & 100.0\% \\ 
\textbf{vector}              & 10 (94.6\%)   & 22          & 22  & 100.0\% \\ 
\midrule
\textbf{Average}             & 9.44 (94.0\%) & 17.22       & 21.78 & 100.0\% \\ 
\bottomrule
\end{tabular}
\caption{For each example, the table shows the number and line coverage of \textit{filtering tests}, the number of \textit{filtered invariants} without refinement, and the number of invariants after 1 refinement attempt. With \textit{filtering tests} and 1 refinement, \tech achieves a perfect unit test pass rate.}
\label{ds_refine}
\end{table}

%% file: z3.tex
\begin{table}[!htp]\centering
\caption{Statistics of the studied data structures in Z3}\label{tab:z3-data}
\small
\begin{tabular}{l|rrrrrrrr}\toprule
&ema &dlist &heap &hashtable &permutation &scoped\_vector &bdd\_manager \\\midrule
\# LoC &57 &243 &309 &761 &177 &220 &1635 \\
\# dependencies &0 &2 &1 &9 &2 &2 &13 \\
\bottomrule
\end{tabular}
\end{table}

As a real-world case study, we apply \tech to synthesize invariants for 7 core data structures from Z3~\cite{z3}, ranging from the simple 57-line \CodeIn{ema} class to the complex 1635-line \CodeIn{bdd\_manager}. The complete set includes \CodeIn{dlist}, \CodeIn{heap}, \CodeIn{hashtable}, \CodeIn{permutation}, and \CodeIn{scoped\_vector}, with varying implementation complexity and the number of dependent classes as shown in Table~\ref{tab:z3-data}. 
Our results were validated by one of the Z3 authors, who confirmed at least one \emph{correct and useful} invariant for each studied class, with the \CodeIn{bdd\_manager} class yielding 11 valuable invariants including the 2 already written by Z3 authors.

Z3 is a widely adopted SMT solver used in a variety of high-stakes applications that depend on its correctness. It is integrated into tools like LLVM~\cite{llvm}, KLEE~\cite{klee}, Dafny~\cite{leino2010dafny} and Frama-C~\cite{kirchner2015frama}.
We selected the Z3 codebase due to its stringent correctness requirements; as an SMT solver, Z3 is employed in applications demanding high reliability. This high-stakes environment makes Z3 an ideal testbed for assessing the effectiveness of synthesized invariants.


The \CodeIn{bdd\_manager} class\footnote{\url{https://github.com/Z3Prover/z3/blob/master/src/math/dd/dd_bdd.h}} is particularly noteworthy. 
It was chosen because it is a self-contained example with developer-written unit tests for validation, presenting a realistic yet manageable challenge. 
Note that the existing developer tests were used after invariants were generated, not as input to the LLM. 
The \CodeIn{bdd\_manager} class in Z3 is a utility for managing Binary Decision Diagrams (BDDs)~\cite{bdd-paper}, which are data structures used to represent Boolean functions efficiently. In BDDs, Boolean functions are represented as directed acyclic graphs, where each non-terminal node corresponds to a Boolean variable, and edges represent the truth values of these variables (\textit{true} or \textit{false}). This representation simplifies complex Boolean expressions and enables efficient operations on Boolean functions.

With 382 lines of code in its header and 1253 lines in the implementation file, \CodeIn{bdd\_manager} surpasses standard data structure complexity, offering an opportunity to evaluate \tech's capability to generate meaningful invariants relevant to real-world scenarios. 
\tech achieves this by compositional generation, recursively traversing the source program's AST. 
Recursive generation became crucial when handling large classes like \CodeIn{bdd\_manager}, which exceeded the LLM’s context window. Decomposing and processing its components separately allowed us to fit relevant parts into the model’s input, demonstrating the utility of recursive invariant generation for large codebases. This supports its relevance in real-world applications beyond the benchmarks.

The \CodeIn{bdd\_manager} class has a developer-written member function for checking its well-formedness, as shown in Figure~\ref{fig:bdd_well_formed}, which we removed during \tech generation. Of the 56 invariants generated by \tech, one of Z3's main authors identified 11 distinct \textit{correct and useful} invariants (e.g., Figure~\ref{fig:bdd_correct_useful}) including the 2 developer-written invariants; these invariants could potentially be integrated into the codebase.
An additional 5 distinct \textit{ok} invariants (e.g., Figure~\ref{fig:bdd_ok})  are labeled correct but have limited utility, 16 distinct \textit{correct but useless} invariants (e.g., those already checked during compilation, such as type checks and constants, Figure~\ref{fig:bdd_correct_useless}), and 2 \textit{incorrect} invariants (e.g., Figure~\ref{fig:bdd_incorrect}). The remaining invariants were repetitions within these categories. This evaluation aligns with \tech's validation results, as our validation pipeline also identified 2 incorrect invariants that failed \CodeIn{bdd\_manager} unit tests.
\shuvendu{Why did you choose to report invariants that already failed the unit test?}

Overall, the Z3 authors' evaluation results further confirm \tech's potential utility in real-world, large-scale codebases.

\begin{figure}[htp]
    \centering
\begin{lstlisting}[language=c++]
bool bdd_manager::well_formed() {
    bool ok = true;
    for (unsigned n : m_free_nodes) {
        ok &= (lo(n) == 0 && hi(n) == 0 && m_nodes[n].m_refcount == 0);
        if (!ok) {
            IF_VERBOSE(0, verbose_stream() << "free node is not internal " << n << " " << lo(n) << " " << hi(n) << " " << m_nodes[n].m_refcount << "\n";
            display(verbose_stream()););
            UNREACHABLE();
            return false;
        }
    }
    
    for (bdd_node const& n : m_nodes) {
        if (n.is_internal()) continue;
        unsigned lvl = n.m_level;
        BDD lo = n.m_lo;
        BDD hi = n.m_hi;
        ok &= is_const(lo) || level(lo) < lvl;
        ok &= is_const(hi) || level(hi) < lvl;
        ok &= is_const(lo) || !m_nodes[lo].is_internal();
        ok &= is_const(hi) || !m_nodes[hi].is_internal();
        if (!ok) {
            IF_VERBOSE(0, display(verbose_stream() << n.m_index << " lo " << lo << " hi " << hi << "\n"););
            UNREACHABLE();
            return false;
        }
    }
    return ok;
}
\end{lstlisting}
    \caption{Z3 developer-written class invariants for \CodeIn{bdd\_manager} class}
    \label{fig:bdd_well_formed}
\end{figure}

\begin{figure}[htp]
    \centering
\begin{lstlisting}[language=c++]
// Node consistency: Each node's index should match its position in m_nodes
for (unsigned i = 0; i < m_nodes.size(); ++i) {
    assert(m_nodes[i].m_index == i);
}
\end{lstlisting}
    \caption{\textit{Correct and useful} invariant for \CodeIn{bdd\_manager} class}
    \label{fig:bdd_correct_useful}
\end{figure}

\begin{figure}[htp]
    \centering
\begin{lstlisting}[language=c++]
// Cache consistency: Entries in the operation cache should be valid
for (const auto* e : m_op_cache) {
    assert(e != nullptr);
    assert(e->m_result != null_bdd);
}
\end{lstlisting}
    \caption{\textit{Ok} invariant for \CodeIn{bdd\_manager} class}
    \label{fig:bdd_ok}
\end{figure}

\begin{figure}[htp]
    \centering
\begin{lstlisting}[language=c++]
// m_is_new_node is a boolean
assert(m_is_new_node == true || m_is_new_node == false);
\end{lstlisting}
    \caption{\textit{Correct and useless} invariant for \CodeIn{bdd\_manager} class}
    \label{fig:bdd_correct_useless}
\end{figure}


\begin{figure}[htp]
    \centering
\begin{lstlisting}[language=c++]
// The number of nodes should not exceed the maximum number of BDD nodes
assert(m_nodes.size() <= m_max_num_bdd_nodes);
\end{lstlisting}
    \caption{\textit{Incorrect} invariant for \CodeIn{bdd\_manager} class}
    \label{fig:bdd_incorrect}
\end{figure}

%% file: related-work.tex

In this section, we discuss how \tech relates to previous works on synthesizing program invariants statically, dynamically, and neurally.



\subsection{Static approaches}
Static techniques, such as interpolation~\cite{mcmillan2004interpolants} or abstract interpretation~\cite{cousot1977abstract} perform a symbolic analysis of source code to compute static over-approximations of runtime behavior and represent them as program invariants over suitable domains.
These techniques are often used to prove the safety properties of the code.
They focus on synthesizing loop invariants and method pre/postconditions, and a few also handle module-level specifications~\cite{lahiri-cav09}. 
Given the undecidability of program verification, these techniques scale poorly for real-world programs, especially in the presence of complex data structures and frameworks. 
In contrast, \tech can be applied to large codebases to synthesize high-quality class invariants but does not guarantee soundness by construction. 

\subsection{Dynamic approaches}
Dynamic synthesis techniques, such as Daikon~\cite{ernst2007daikon}, DIG~\cite{nguyen2012dig}, SLING~\cite{le2019sling}, and specification mining~\cite{ammons2002}, learn invariants by observing the dynamic behaviors of programs over a set of concrete execution traces. 
One advantage of these dynamic techniques is that they can be agnostic to the code and generally applicable to different languages. 
However, these approaches are limited by the templates or patterns over which the invariants can be expressed. 
DySy~\cite{dysy} employs dynamic symbolic execution to alleviate the problem of fixed templates for bounded executions but resorts to ad-hoc abstraction for loops or recursion. 
\citep{hellendoorn2019are} trained models to predict the quality of invariants generated by tools such as Daikon, but they do not generate new invariants. 
SpecFuzzer~\cite{facundo2022specfuzzer} generates numerous candidate assertions via fuzzing to construct templates and filters them using Daikon and mutation testing. 
Finally, Geminus~\citep{boockmann2024geminius} aims at synthesizing sound and complete class invariants representing the set of reachable states, guiding their search using random test cases termed Random Walk.

Unlike these approaches, \tech can generate a much larger class of invariants, leveraging multimodal inputs, including source code, test cases, comments, and even the naming convention learned from training data, to enhance invariant synthesis.
Further, unlike prior dynamic approaches, LLM-based test generation (an active area of research~\cite{codamosa-icse23,schäfer2023empiricalevaluationusinglarge,yang2024whitefox}) reduces the need to have a high-quality test suite to obtain the invariants.

For the use case of static verification, learning-based approaches have been used to iteratively improve the quality of the synthesized inductive invariants~\cite{garg2014learning, garg2016learning, padhi2016loopinvgen} from dynamic traces. 
However, these approaches have not been evaluated in real-world programs due to the need for symbolic reasoning. 

\subsection{Neural approaches}
LLM-based invariant synthesis is an emerging area of research with some noteworthy recent contributions. \citep{pei2023learning} trained a model for zero-shot invariant synthesis, which incurs high training costs and lacks feedback-driven repair. 
Their approach uses Daikon-generated invariants as both training data and ground truth, which can lead to spurious invariants. 

Prior work on nl2postcond~\cite{nl2postcond} prompts LLMs to generate pre- and postconditions of Python and Java benchmarks, illustrating LLMs' ability to generate high-quality specifications. 
However, they do not prune incorrect invariants and do not generate class invariants. 
Combine this work with \tech would be an interesting direction for future work.

Two very recent neuro–symbolic pipelines extend LLM prompting to \emph{other} kinds of specifications. \cite{WuASE24} combines GPT-4 with bounded-model checking to infer \emph{loop invariants}: the LLM enumerates candidate predicates, a BMC oracle filters them, and the surviving predicates are re-assembled into provable invariants, yielding a 97\,\% success rate on 316 numeric-loop benchmarks.
\cite{WenCAV24} (\textsc{AutoSpec}) weaves static slicing and an off-the-shelf program verifier with LLM generation to synthesise \emph{function-level contracts}; AutoSpec verifies 79\,\% of heterogeneous benchmarks plus an X.509 parser case study.
Both systems rely on \emph{static} or SMT-based oracles and target scalar loops or procedure specifications,
whereas \tech tackles \emph{pointer-rich class/object invariants} in idiomatic~C++ and validates them chiefly through \emph{dynamic} test-suite execution plus mutation testing.
The different oracle allows our approach to scale to data-structure code bases where precise SMT models are hard to obtain.

For static verification, recent works include the use of LLM for intent formalization from natural language~\cite{lahiri2024evaluating},  and inferring specifications and inductive program invariants~\cite{loopy,ma2024specgen}.
None of these techniques scale to real-world programs due to the need for complex symbolic reasoning.

%% file: discussion.tex

At present, \tech judges an invariant’s correctness with the same test suite that the LLM co-generates alongside that invariant. This design keeps the pipeline fully automated, but it also risks co-adaptation: the model can drift toward invariants that merely fit the behaviours exercised by its own tests, overstating their generality.
\tech uses generated tests for invariant pruning, but the test suite may include spurious tests that can incorrectly prune valid invariants. The generated tests might not represent valid sequences of method calls; for example, invoking a \CodeIn{pop()} method before a \CodeIn{push()} method could fail certain assertions, leading to improper pruning.

Another limitation is the LLM's context window, which restricts the amount of code that can be processed in a single call. This limitation makes it challenging to handle large codebases. \tech partially addresses this issue through compositional generation, breaking down the code into manageable parts. Ongoing advancements in LLMs, as highlighted in recent work~\cite{liu2024lost,gao2023retrieval}, are also expected to mitigate this limitation.

For future work, we plan to integrate invariant generation with the generation of formal specifications for member functions, enabling LLMs to gain a more comprehensive understanding of program behavior. Additionally, we aim to evaluate \tech on larger and more complex systems beyond Z3, demonstrating its scalability to diverse codebases.






%% file: concl.tex
In this paper, we describe an approach to leverage LLMs and a lightweight mixed static/dynamic approach to synthesize class invariants. 
Our experiments on standard C++ data structures as well as a popular and high-assurance codebase demonstrate the feasibility of our approach.
Our technique is currently limited by an automated way to integrate the generated tests into the build system of the underlying repo, and the need for developers to validate the invariants.
We envision that integrating \tech with continuous integration (CI) and pull requests (PR) can aid in scaling the approach to more developers.
In future work, we also plan to investigate incorporating developer feedback to repair or strengthen generated invariants.

%% file: prompt_refine.tex
\section{Prompt}
\label{appendix:prompt}
\begin{figure*}[htp]
\begin{lstlisting}[language=markdown]
You are an expert in creating program invariants from code and natural language.
Invariants are assertions on the variables in scope that hold true at different program points
We are interested in finding invariants that hold at both start and end of a function within a data structure. Such an invariant is commonly known as an object invariant.  

The invariants can usually be expressed as a check on the state at the particular program point. The check should be expressed as a check in the same underlying programming language which evaluates to true or false. To express these, you can use:
- An assertion in the programming language
- A pure method (which does not have any side effect on the variables in scope) that checks one or more assertion
- For a collection, you can use a loop to iterate over elements of the collection and assert something on each element or a pair of elements.  

Task Description: 
Task 1: Given a module, in the form of a class definition, your task is to infer object invariants about the class. For doing so, you may examine how the methods of the class read and modify the various fields of the class. 
For coming up with invariants, you may use the provided code and any comments in the code. You may also use world knowledge to guide the search for invariants. 
 
Task 2: Generate unit tests for the class based on the class definition and public API methods. The test cases should simulate a series of public method calls to verify the behavior of the class, but do not use any testing framework like gtest. Do not add `assert` or any form of assertions.
\end{lstlisting}
    \caption{\tech Generation system prompt: instruction and task description. Lines 3-7 give the formal definition of a class invariant (must hold before and after every public method).}
    \label{fig:prompt_generation_system_task_app}
\end{figure*}

\begin{figure}
    \centering
    \begin{lstlisting}[language=markdown,firstnumber=15]
Input Format:
You will be given the name of a class or typedef, and a section of code containing the definition of the class. You will also be given the definitions of functions that read and modify the fields of the class. 

Output Format:
The output should be in the following format:

The first paragraph should begin with "REASONING:". From the next line onwards, it should contain the detailed reasoning and analysis used for the inference of the object invariants. The entire text should be enclosed in $$$. For example,
``$$$
REASONING: 
explanation
$$$

The next paragraph should begin with "INVARIANTS:". From the next line onwards, it should contain a list of the various invariants inferred. The invariants should be in the form of code in the same underlying programming language, enclosed by ```. Each invariant should start from a new line, and be separated by "---". Use lambda if necessary. If lambda is recursive, explicitly specify the type of the lambda function and use `std::function` for recursion. Do not use helper functions. For example, 
INVARIANTS: 
```/* Invariant 1 */```
---
```/* Multi line Invariant 2  */
    assert(...); ```
---
```/* Invariant 3 */```

The next paragraph should begin with "TESTS:". From the next line onwards, it should contain a list of a API call sequence in the form of code enclosed by ```. Each test should start from a new line, and be separated by "---". For example, 
TESTS: 
```/* Test 1 */
   this->method1();
   this->method2();```
---
```/* Test 2 */
    this.method3(...); ```
---
```/* Test 3 */```

Important:
1. Follow the output format strictly, particularly enclosing each invariant in triple-ticks (```), and enclosing the reasoning in $$$.
2. Only find object invariants for the target class provided to you, do not infer invariants for any other class. 
3. Make sure the invariant is a statement in the same underlying programming language as the source program.
4. If you can decompose a single invariant into smaller ones, try to output multiple invariants.
\end{lstlisting}
\vspace{-0.1in}
    \caption{\tech Generation system prompt: input-output format.}
    \label{fig:prompt_generation_system_inputoutput}
\end{figure}

Figure~\ref{fig:prompt_generation_system_task_app} and Figure~\ref{fig:prompt_generation_system_inputoutput} show the system prompt used by \tech for invariant-test co-generation. The former presents the instruction and task description, while the latter illustrates the input-output format.

\begin{figure*}[htp]
\begin{lstlisting}[language=markdown]
Name of Data Structure to Annotate: {struct}
Code:
```
{code}
```
\end{lstlisting}
\vspace{-0.1in}
    \caption{\tech Generation user prompt template.}
    \label{appendix:prompt_generation_user}
\end{figure*}

\begin{figure*}[htp]
\begin{lstlisting}[language=markdown]
You are an expert in repairing program invariants from code and natural language.
Invariants are assertions on the variables in scope that hold true at different program points. 

We are interested in finding invariants that hold at both start and end of a function within a data structure. Such an invariant is commonly known as an object invariant.  

The invariants can usually be expressed as a check on the state at the particular program point. The check should be expressed as a check in the same underlying programming language which evaluates to true or false. To express these, you can use:
- An assertion in the programming language
- A pure method (which does not have any side effect on the variables in scope) that checks one or more assertion
- For a collection, you can use a loop to iterate over elements of the collection and assert something on each element or a pair of elements.  

Task Description: 
Given a module, in the form of a class definition, your task is to infer object invariants about the class. For doing so, you may examine how the methods of the class read and modify the various fields of the class. 

For coming up with invariants, you may use the provided code and any comments in the code. You may also use world knowledge to guide the search for invariants. 

Input Format:
You will be given the name of a class or typedef, and a section of code containing the definition of the class. You will also be given the definitions of functions which read and modify the fields of the class. 

Output Format:
The output should be in the following format:

The first paragraph should begin with "REASONING:". From the next line onwards, it should contain the detailed reasoning and analysis used for the inference of the object invariants. The entire text should be enclosed in $$$. For example,
``$$$
REASONING: 
explanation
$$$``

The next paragraph should begin with "INVARIANTS:". From the next line onwards, it should contain a list of the various invariants inferred. The invariants should be in the form of code in the same underlying programming language, enclosed by ```. Each invariant should start from a new line, and be separated by "---". For example, 
INVARIANTS: 
```/* Invariant 1 */```
---
```/* Multi line Invariant 2  */
    assert(...);```
---
```/* Invariant 3 */```

Important:
1. Follow the output format strictly, particularly enclosing each invariant in triple-ticks (```), and enclosing the reasoning in $$$.
2. Only find object invariants for the target class provided to you, do not infer invariants for any other class. 
3. Make sure the invariant is a statement in the same underlying programming language as the source program.
4. If you can decompose a single invariant into smaller ones, try to output multiple invariants.
\end{lstlisting}
    \caption{\tech Refinement system prompt: instruction and task description.}
    \label{appendix:prompt_refinement_system}
\end{figure*}

\begin{figure*}[htp]
\begin{lstlisting}[language=markdown]
Please fix the failed invariants given the feedback, tests and the original source code.

Failed Invariant:
```
{invariant}
```

Error message for the failed invariant:
```
{feedback}
```

Name of Data Structure to Annotate: {struct}

Original Code:
```
{code}
```

Gold Tests that Fail the Invariant:
{tests}

\end{lstlisting}
\vspace{-0.1in}
    \caption{\tech Refinement user prompt template.}
    \label{appendix:prompt_refinement_user}
\end{figure*}

%% file: appendix/inv_table.tex
\section{Daikon Invariants Frequency Tables}
\label{appendix:daikon}

\begin{table}[ht]
\centering
\scriptsize
\caption{Invariants for avl\_tree, 11 public methods.}
\label{avl_tree_daikon}
\begin{tabular}{|l|c|}
\hline
Invariant & Count \\
\hline
this->root has only one value & 4 \\
this->root.\_M\_t has only one value & 4 \\
this->root.\_M\_t.\_\_uniq\_ptr\_impl<AvlTree::Node, std::default\_delete<AvlTree::Node> >.\_M\_t has only one value & 4 \\
this[0] has only one value & 3 \\
this->n one of \{ 3, 4 \} & 3 \\
this[0] != null & 2 \\
this->root != null & 2 \\
this->root.\_M\_t != null & 2 \\
this->root.\_M\_t.\_\_uniq\_ptr\_impl<AvlTree::Node, std::default\_delete<AvlTree::Node> >.\_M\_t != null & 2 \\
this->n one of \{ 1, 2, 3 \} & 1 \\
this->n one of \{ 0, 3 \} & 1 \\
this->n >= 1 & 1 \\
this->n == 3 & 1 \\
t.root has only one value & 1 \\
t.root.\_M\_t has only one value & 1 \\
t.root.\_M\_t.\_\_uniq\_ptr\_impl<AvlTree::Node, std::default\_delete<AvlTree::Node> >.\_M\_t has only one value & 1 \\
t.n == 3 & 1 \\
this->n == return & 1 \\
\hline
\end{tabular}
\end{table}

\begin{table}[ht]
\centering
\scriptsize
\caption{Invariants for red\_black\_tree, 11 public methods.}
\label{red_black_tree_daikon}
\begin{tabular}{|l|c|}
\hline
Invariant & Count \\
\hline
this[0] != null & 3 \\
this->root != null & 3 \\
this->root.\_M\_t != null & 3 \\
this->root.\_M\_t.\_\_uniq\_ptr\_impl<RedBlackTree::Node, std::default\_delete<RedBlackTree::Node> >.\_M\_t != null & 3 \\
this->n one of \{ 3, 4, 6 \} & 3 \\
this->root has only one value & 3 \\
this->root.\_M\_t has only one value & 3 \\
this->root.\_M\_t.\_\_uniq\_ptr\_impl<RedBlackTree::Node, std::default\_delete<RedBlackTree::Node> >.\_M\_t has only one value & 3 \\
this[0] has only one value & 2 \\
this->n >= 0 & 1 \\
(No intersection exists) & 1 \\
t.root has only one value & 1 \\
t.root.\_M\_t has only one value & 1 \\
t.root.\_M\_t.\_\_uniq\_ptr\_impl<RedBlackTree::Node, std::default\_delete<RedBlackTree::Node> >.\_M\_t has only one value & 1 \\
t.n == 3 & 1 \\
No intersection & 1 \\
\hline
\end{tabular}
\end{table}

\begin{table}[ht]
\centering
\scriptsize
\caption{Invariants for linked\_list, 8 public methods.}
\label{linked_list_daikon}
\begin{tabular}{|l|c|}
\hline
Invariant & Count \\
\hline
this[0] has only one value & 5 \\
this->head has only one value & 5 \\
this->head.\_M\_t has only one value & 5 \\
this->head.\_M\_t.\_\_uniq\_ptr\_impl<LinkedList::Node, std::default\_delete<LinkedList::Node> >.\_M\_t has only one value & 5 \\
this[0] != null & 4 \\
this->head != null & 4 \\
this->head.\_M\_t != null & 4 \\
this->head.\_M\_t.\_\_uniq\_ptr\_impl<LinkedList::Node, std::default\_delete<LinkedList::Node> >.\_M\_t != null & 4 \\
this->tail[] elements != null & 2 \\
this->tail[].next elements != null & 2 \\
this->tail[].next.\_M\_t elements != null & 2 \\
this->n >= 0 & 2 \\
this->n == 0 & 1 \\
this->tail[].data elements one of \{ 1 \} & 1 \\
this->tail[].data one of \{ [1] \} & 1 \\
this->n one of \{ 1 \} & 1 \\
this->n one of \{ 1, 2 \} & 1 \\
this->tail[].data elements >= 1 & 1 \\
this->tail[].data elements <= this->n & 1 \\
this->tail[].data elements one of \{ 1, 2, 4 \} & 1 \\
this->tail[].data one of \{ [1], [2], [4] \} & 1 \\
this->n one of \{ 2, 3 \} & 1 \\
this->tail[].data == [3] & 1 \\
this->tail[].data elements == 3 & 1 \\
\hline
\end{tabular}
\end{table}

\begin{table}[ht]
\centering
\scriptsize
\caption{Invariants for binary\_search\_tree, 11 public methods.}
\label{binary_search_tree_daikon}
\begin{tabular}{|l|c|}
\hline
Invariant & Count \\
\hline
this->root has only one value & 4 \\
this->root.\_M\_t has only one value & 4 \\
this->root.\_M\_t.\_\_uniq\_ptr\_impl<BinarySearchTree::Node, std::default\_delete<BinarySearchTree::Node> >.\_M\_t has only one value & 4 \\
this->n one of \{ 0, 2, 3 \} & 3 \\
this->n one of \{ 0, 3 \} & 3 \\
this[0] has only one value & 3 \\
this[0] != null & 2 \\
this->root != null & 2 \\
this->root.\_M\_t != null & 2 \\
this->root.\_M\_t.\_\_uniq\_ptr\_impl<BinarySearchTree::Node, std::default\_delete<BinarySearchTree::Node> >.\_M\_t != null & 2 \\
t.root has only one value & 1 \\
t.root.\_M\_t has only one value & 1 \\
t.root.\_M\_t.\_\_uniq\_ptr\_impl<BinarySearchTree::Node, std::default\_delete<BinarySearchTree::Node> >.\_M\_t has only one value & 1 \\
t.n == 3 & 1 \\
(this->n == return) == (return == orig(this->n)) & 1 \\
\hline
\end{tabular}
\end{table}

\begin{table}[ht]
\centering
\scriptsize
\caption{Invariants for heap, 7 public methods.}
\label{heap_daikon}
\begin{tabular}{|l|c|}
\hline
Invariant & Count \\
\hline
this->comp.\_M\_invoker has only one value & 7 \\
this->comp.\_Function\_base.\_M\_manager has only one value & 7 \\
this[0] has only one value & 6 \\
this->data has only one value & 6 \\
this->data.\_Vector\_base<int, std::allocator<int> >.\_M\_impl has only one value & 6 \\
this->comp has only one value & 6 \\
this->comp.\_Function\_base.\_M\_functor has only one value & 6 \\
this->comp.\_Function\_base.\_M\_functor.\_M\_unused has only one value & 6 \\
this[0] != null & 3 \\
this->data != null & 3 \\
this->data.\_Vector\_base<int, std::allocator<int> >.\_M\_impl != null & 3 \\
this->comp != null & 3 \\
this->comp.\_M\_invoker != null & 3 \\
this->comp.\_Function\_base.\_M\_functor != null & 3 \\
this->comp.\_Function\_base.\_M\_functor.\_M\_unused != null & 3 \\
this->comp.\_Function\_base.\_M\_manager != null & 3 \\
this->data.\_Vector\_base<int, std::allocator<int> >.\_M\_impl.\_Vector\_impl\_data.\_M\_start[] elements >= 1 & 2 \\
this->data.\_Vector\_base<int, std::allocator<int> >.\_M\_impl.\_Vector\_impl\_data.\_M\_start != null & 1 \\
this->data.\_Vector\_base<int, std::allocator<int> >.\_M\_impl.\_Vector\_impl\_data.\_M\_finish != null & 1 \\
this->data.\_Vector\_base<int, std::allocator<int> >.\_M\_impl.\_Vector\_impl\_data.\_M\_end\_of\_storage != null & 1 \\
\hline
\end{tabular}
\end{table}

\begin{table}[ht]
\centering
\scriptsize
\caption{Invariants for hash\_table, 7 public methods.}
\label{hash_table_daikon}
\begin{tabular}{|l|c|}
\hline
Invariant & Count \\
\hline
::\_\_digits == "000102...6979899" & 3 \\
::\_\_tag == "" & 3 \\
this[0] has only one value & 3 \\
this->hash\_function has only one value & 3 \\
this->hash\_function.\_Function\_base.\_M\_functor has only one value & 3 \\
this->hash\_function.\_Function\_base.\_M\_functor.\_M\_unused has only one value & 3 \\
this->hash\_function.\_Function\_base.\_M\_functor.\_M\_pod\_data == "" & 3 \\
this->load\_factor == 0.75 & 3 \\
this->table has only one value & 3 \\
this->table.\_Vector\_base<... >.\_M\_impl has only one value & 2 \\
this->table.\_Vector\_base<... >.\_M\_impl has only one value & 1 \\
this->\_num\_elements one of \{ 0, 1 \} & 1 \\
this->\_size == 10 & 1 \\
this->table.\_Vector\_base<... > > > > >.\_M\_impl.\_Vector\_impl\_data.\_M\_end\_of\_storage & 1 \\
key.\_M\_dataplus has only one value & 1 \\
key.\_M\_dataplus.\_M\_p one of \{ "key1", "key2" \} & 1 \\
key.\_M\_string\_length == 4 & 1 \\
this->\_num\_elements >= 0 & 1 \\
this->\_size >= 0 & 1 \\
this->\_num\_elements <= this->\_size & 1 \\
\hline
\end{tabular}
\end{table}

\begin{table}[ht]
\centering
\scriptsize
\caption{Invariants for vector, 11 public methods. }
\label{vector_daikon}
\begin{tabular}{|l|c|}
\hline
Invariant & Count \\
\hline
this[0] has only one value & 9 \\
this->capacity == 5 & 7 \\
this->data has only one value & 5 \\
this->data[] elements one of \{ 1, 2 \} & 5 \\
this->n == 2 & 4 \\
this->data[] == [1, 2] & 4 \\
this->n in this->data[] & 4 \\
this->data[] == [1] & 3 \\
this->n == 0 & 2 \\
this->n one of \{ 1 \} & 2 \\
this->n in return[] & 1 \\
return[] == [1, 2] & 1 \\
return[] elements one of \{ 1, 2 \} & 1 \\
this->capacity == 0 & 1 \\
this->data == null & 1 \\
this->n == 1 & 1 \\
this->n == v.n & 1 \\
this->capacity == v.capacity & 1 \\
v.data has only one value & 1 \\
v one of \{ 1, 2 \} & 1 \\
this->capacity one of \{ 5 \} & 1 \\
this->data[] sorted by < & 1 \\
this->n <= this->capacity & 1 \\
this->n < this->capacity & 1 \\
this->n one of \{ 2, 5 \} & 1 \\
this->data[] elements == 1 & 1 \\
this->data[] one of \{ [1], [1, 2] \} & 1 \\
return one of \{ 1, 2 \} & 1 \\
\hline
\end{tabular}
\end{table}

\begin{table}[ht]
\centering
\scriptsize
\caption{Invariants for queue, 7 public methods.}
\label{queue_daikon}
\begin{tabular}{|l|c|}
\hline
Invariant & Count \\
\hline
this->data has only one value & 5 \\
this->maxSize == 10 & 5 \\
this[0] has only one value & 4 \\
this->data[] == [10, 20, 30] & 2 \\
this->data[] elements one of \{ 10, 20, 30 \} & 2 \\
this->head one of \{ 0, 1 \} & 2 \\
this->tail == 3 & 2 \\
this->n one of \{ 2, 3 \} & 2 \\
this->maxSize in this->data[] & 2 \\
this[0] != null & 2 \\
this->data != null & 2 \\
this->data[] elements >= 0 & 2 \\
this->data[] sorted by < & 2 \\
this->head < this->maxSize & 2 \\
this->tail < this->maxSize & 2 \\
this->data[] elements one of \{ 1, 2 \} & 2 \\
this->data[] one of \{ [1], [1, 2] \} & 2 \\
this->tail one of \{ 0, 1, 2 \} & 2 \\
this->tail in this->data[] & 2 \\
this->head - this->tail + this->n == 0 & 1 \\
this->tail one of \{ 2, 3, 100 \} & 1 \\
this->maxSize one of \{ 10, 160 \} & 1 \\
this->n < this->maxSize & 1 \\
this->head one of \{ 0, 50 \} & 1 \\
this->tail >= 0 & 1 \\
this->n <= this->maxSize & 1 \\
this->head <= this->tail & 1 \\
this->head one of \{ 0, 2 \} & 1 \\
this->n one of \{ 0, 1 \} & 1 \\
this->head == other.head & 1 \\
this->maxSize == other.maxSize & 1 \\
this->data[] == [7, 14] & 1 \\
this->data[] elements one of \{ 7, 14 \} & 1 \\
this->head == 0 & 1 \\
other.data has only one value & 1 \\
other.tail == 2 & 1 \\
this->head one of \{ 0, 1, 2 \} & 1 \\
this->head - this->tail + return == 0 & 1 \\
return one of \{ 0, 1, 2 \} & 1 \\
\hline
\end{tabular}
\end{table}

\begin{table}[ht]
\centering
\scriptsize
\caption{Invariants for stack, 6 public methods.}
\label{stack_daikon}
\begin{tabular}{|l|c|}
\hline
Invariant & Count \\
\hline
this->maxSize one of \{ 10, 160, 1280 \} & 4 \\
this->data[] sorted by < & 4 \\
this->data[] elements < this->maxSize & 4 \\
this->n < this->maxSize & 3 \\
this[0] != null & 3 \\
this->data != null & 3 \\
this[0] has only one value & 2 \\
this->data[] elements >= 0 & 2 \\
this->n one of \{ 0, 1, 2 \} & 1 \\
this->maxSize == other.maxSize & 1 \\
this has only one value & 1 \\
this->data has only one value & 1 \\
this->n == 2 & 1 \\
this->maxSize == 10 & 1 \\
other.data has only one value & 1 \\
other.data[] == [1, 2] & 1 \\
other.data[] elements one of \{ 1, 2 \} & 1 \\
this->n <= this->maxSize & 1 \\
\hline
\end{tabular}
\end{table}

%% file: appendix/approach_details.tex
\section{Additional Implementation Details}
\label{sec:appendix_approach_details}

This appendix provides additional details and examples for the implementation of \tech.

\subsection{Code Instrumentation and Invariant Examples}
\label{subsec:appendix_instrumentation}

Figure \ref{fig:appendix_avl_instrumentation_combined} shows how a public method is instrumented with invariant checks, and Figure \ref{fig:appendix_avl_incorrect} shows an example of an incorrect invariant.

\begin{figure}[ht]
\centering
\begin{subfigure}[t]{0.45\linewidth}
\begin{lstlisting}[language=c++, escapechar=@, basicstyle=\ttfamily\scriptsize]
bool AvlTree::empty() { 
  check_invariant();
  auto ret = empty_original();
  check_invariant();
  return ret;
}

bool AvlTree::empty_original() { 
  return n == 0; 
}
\end{lstlisting}
\caption{AvlTree instrumented with invariants}
\label{fig:appendix_avl_instrumentation}
\end{subfigure}
\hfill
\begin{subfigure}[t]{0.53\linewidth}
\begin{lstlisting}[language=c++, escapechar=@, basicstyle=\ttfamily\scriptsize]
void AvlTree::check_invariant() {
  std::function<bool(const std::unique_ptr<Node>&)> 
  is_balanced = [&](const std::unique_ptr<Node>& node) -> bool {
    if (!node) return true;
    int left_height = height(node->left);
    int right_height = height(node->right);
    if (std::abs(left_height - right_height) > 1) 
        return false;
    return is_balanced(node->left) && 
           is_balanced(node->right);
  };
  assert(is_balanced(root));
}
\end{lstlisting}
\caption{Example of a correct AvlTree class invariant}
\label{fig:appendix_avl_correct_inv2}
\end{subfigure}
\caption{AvlTree instrumentation and invariant example}
\label{fig:appendix_avl_instrumentation_combined}
\end{figure}

\begin{figure}[ht]
\centering
\begin{lstlisting}[language=c++, escapechar=@, basicstyle=\ttfamily\scriptsize]
void AvlTree::check_invariant() {
    assert(height(root) == get_height(root));
}
\end{lstlisting}
\caption{Example of an incorrect invariant of AvlTree because there is no get\_height method}
\label{fig:appendix_avl_incorrect}
\end{figure}

\subsection{Test Generation Examples}
\label{subsec:appendix_test_generation}

Figure \ref{fig:appendix_avl_tests} shows an example of a generated test suite used for filtering invariants.

\begin{figure}[htp]
\centering
\begin{lstlisting}[language=c++, escapechar=!, basicstyle=\ttfamily\scriptsize]
int main() {
    // Test Case 1: Basic insertions and traversals
    {
        AvlTree tree;
        tree.insert(10); tree.insert(20); tree.insert(5);
        tree.in_order_traversal();
        tree.pre_order_traversal();
    }
    // Test Case 2: Size, height, empty checks
    {
        AvlTree tree;
        tree.insert(10); tree.insert(20);
        tree.size(); tree.height(); tree.empty();
    }
    // ... more test cases ...
}
\end{lstlisting}
\caption{A test suite generated for AvlTree}
\label{fig:appendix_avl_tests}
\end{figure}

\subsection{Refinement Details}
\label{subsec:appendix_refinement}

\tech implements a feedback loop for refining failing invariants. Figure \ref{fig:appendix_gcc_compiler_error} shows an error message, while Figures \ref{fig:appendix_avl_bst_before} and \ref{fig:appendix_avl_bst_after} show the BST invariant before and after refinement.

\begin{figure}[ht]
\centering
\begin{subfigure}[t]{0.45\linewidth}
\begin{lstlisting}[language=c++, escapechar=@, basicstyle=\ttfamily\scriptsize]
avl_tree.cpp: In lambda function:
avl_tree.cpp:21:16: error: use of 
'is_balanced' before deduction of 'auto'
\end{lstlisting}
\caption{gcc compiler error messages}
\label{fig:appendix_gcc_compiler_error}
\end{subfigure}
\hfill
\begin{subfigure}[t]{0.52\linewidth}
\begin{lstlisting}[language=c++, escapechar=@, basicstyle=\ttfamily\scriptsize]
void AvlTree::check_invariant() {
@\CodeBG@   auto is_bst = [&](const std::unique_ptr<Node>& node, 
  const T& min, const T& max) -> bool {
    if (!node) return true;
    if (node->data <= min || 
        node->data >= max) return false;
    return is_bst(node->left, min, node->data) && 
           is_bst(node->right, node->data, max);
  };
  assert(is_bst(root, std::numeric_limits<T>::min(), 
         std::numeric_limits<T>::max()));
}
\end{lstlisting}
\caption{BST property before refinement}
\label{fig:appendix_avl_bst_before}
\end{subfigure}
\end{figure}

\begin{figure}[ht]
\centering
\begin{lstlisting}[language=c++, escapechar=@, basicstyle=\ttfamily\scriptsize]
void AvlTree::check_invariant() {
@\CodeBGGreen@  std::function<bool(const std::unique_ptr<Node>&, const T&, const T&)> 
  is_bst = [&](const std::unique_ptr<Node>& node, const T& min, const T& max) -> bool {
    if (!node) return true;
    if (node->data <= min || node->data >= max) return false;
    return is_bst(node->left, min, node->data) && 
           is_bst(node->right, node->data, max);
  };
  assert(is_bst(root, std::numeric_limits<T>::min(), std::numeric_limits<T>::max()));
}
\end{lstlisting}
\caption{BST property after refinement: \CodeIn{auto} is changed to explicit declarations}
\label{fig:appendix_avl_bst_after}
\end{figure} 